# F2SD: A dataset for end-to-end group detection algorithms

Giang Hoang[a], Tuan Nguyen Dinh[a], Tung Cao Hoang[a], Son Le Duy[a], Keisuke Hihara[b], Yumeka Utada[b], Akihiko Torii[b], Naoki Izumi[b], and Long Tran Quoc[a]

[a]VNU University of Engineering and Technology, 144 Xuan Thuy, Cau Giay, Hanoi, Vietnam
[b]Dai Nippon Printing Co., Ltd., Japan

## ABSTRACT

The lack of large-scale datasets has been impeding the advance of deep learning approaches to the problem of F-formation detection. Moreover, most research works on this problem rely on input sensor signals of object location and orientation rather than image signals. To address this, we develop a new, large-scale dataset of simulated images for F-formation detection, called **F-f**ormation **S**imulation **D**ataset (*F2SD*). *F2SD* contains nearly 60,000 images simulated from GTA-5, with bounding boxes and orientation information on images, making it useful for a wide variety of modelling approaches. It is also closer to practical scenarios, where three-dimensional location and orientation information are costly to record. It is challenging to construct such a large-scale simulated dataset while keeping it realistic. Furthermore, the available research utilizes conventional methods to detect groups. They do not detect groups directly from the image. In this work, we propose (1) a large-scale simulation dataset *F2SD* and a pipeline for F-formation simulation, (2) a first-ever end-to-end baseline model for the task, and experiments on our simulation dataset.

**Keywords:** group conversations, f-formations, deep neural network

## 1. INTRODUCTION

Humans arrange themselves spatially in various kinds of focused interaction, especially conversation. A. Kendon [1] describes F-Formations as the spatial and orientation relationship between two or more people conversing with each other. Face-to-face conversation groups (FCGs) are the most common form of social interaction, denoting small groups of two or more co-existing persons engaged in ad-hoc interactions [2]. It appears naturally in a variety of social situations and can be represented as a pattern of F-Formation [3], based on this information, robots can understand the environment and interact smoothly with people. Automated analysis of such patterns remains a critical challenge for robotics systems.To address the challenges of analyzing groups, a variety of benchmarks like SALSA [4], Cocktail Party [5] has been presented by researchers in the community. However, available datasets have significant drawbacks. Firstly, the quantity of the data is very limited (usually about a few thousand frames), making it very challenging to conduct experiments with modern methods such as deep learning. Secondly, the highly complex data creation processes usually involve recording a controlled social interaction event. Experts, i.e. psychologists, then annotate the labels. Most current group detection methods recognize groups based on the head pose, body pose and location in two-dimensional coordination. We argue that this pipeline is not convenient. We can employ deep learning methods to discover groups and people directly from the image. Available datasets such as [4] do not contain information on bounding boxes, and there is a lack of variety in condition, camera setup, people and quantity.

Furthermore, various approaches have been made toward the F-formation detection task, ranging from traditional [6] [7] to data-driven methods [8] [9]. The major drawbacks of these works are (1) the need to use additional devices (sensors, depth camera,...) to capture intermediary information like body pose or head orientation or (2) the reliance on heuristically determined threshold values. Most methods rely on hard-coded parameters whose values are derived by experiments or conventional methods from small-scale datasets, which may not be generalizable. For example, [8], [9] are experimented on SALSA dataset [4], [5], etc. These datasets are fixed at the number of people in the frame and camera setup as well as the environment.

We address these problems and propose two novel contributions. First, we propose a large-scale simulated dataset using the open-world game Grand Theft Auto V, with the synthesis group algorithm being based on the distribution of the SALSA dataset. Our proposed dataset contains over 40,000 images for the training set and 20,000 images for the test set with the various conditional environment, number of people, human models and camera configurations. Figure 1 shows some samples of our simulation dataset.

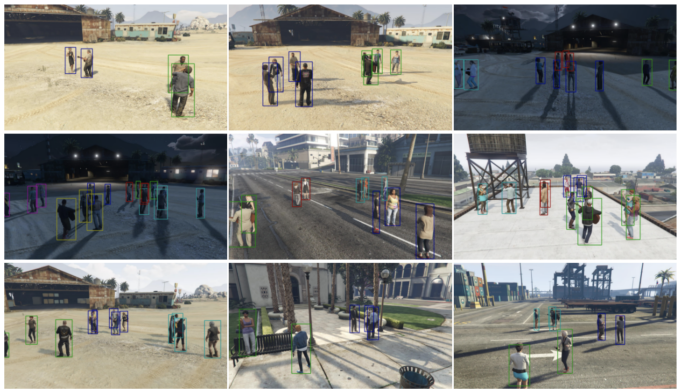

Figure 1: Sample of our simulation dataset

Second, to our knowledge, there are no existing large-scale datasets available to efficiently benchmark the task on modern solutions, nor any methods that detect F-formation directly from visual RGB images. Our proposed method focuses on a vision-based approach, with a large-scale simulated dataset and an end-to-end model that captures F-formation groups directly from input images without intermediary pose estimation or any special setups.

## 2. RELATED WORK

### 2.1 Dataset

The study of modelling F-formation has caught the attention of the computer vision community. As social robots are becoming more common in homes and workplaces, these systems must promote conditions for good social interaction by exploiting the natural spatial interactions that humans use [10].

Some existing works tackle the group FGCs analysis problem by creating real-life datasets. Hung et al.[11], Zen et al.[5], Cristani et al.[12], and Farenzena et al.[13] created visual datasets for F-formation detection. The data creation pipeline in those researches usually involves recording a controlled social interaction event, and experts (i.e. psychologists) annotate the labels. Such setup requires remarkable workloads, thus limiting the ability to scale up the quantity of the samples. Alameda-Pineda et al.[4] addressed the problems of the existing datasets with multi-modal setups, annotated over a long duration and challenging scenes. However, the number of subjects is fixed to 18, thus limiting the generalization to real-life scenarios. Furthermore, although the quantity of annotated frames is larger than in existing works, it is still challenging to conduct experiments with modern methods such as deep learning, which requires an even more significant amount of data.

### 2.2 Method

A wide range of algorithms has been developed for F-formation detection. Together with traditional approaches such as hough voting strategy [6], dominant set [7] or graph cut [14], recent works on F-formation detection have shown a significant improvement in data-driven approaches. Mason et al.[8] built an interaction graph based on top-view spatial features. Each node of the graph corresponds to a social agent (a person or a robot), while undirected edges connect pairs of nodes in the graph. The weight of each edge represents the likelihood that the two connected vertices are in the same group conversation - or the same F-formation cluster. A deep affinity network predicts the graph's affinity matrix, and the Dominant Sets algorithm is then used to find F-formation clusters. Meanwhile, Hedayati et al. [9] proposed a similar method, introducing Distance and Effort Angle as new input features. The problem formulation also involves building a graph to represent the connections between the agents, and the affinity matrix is predicted using Weighted KNN, Bagged Trees and Logistic Regression. The final F-formation groups are constructed using a greedy reconstruction algorithm.

While being very promising, these data-driven methods involve pose estimation to construct the input. This requires the annotated datasets to include head/body orientation information, thus limiting the ability to scale up in quantity.

We propose *F2SD* dataset and a novel end-to-end model to address the problem of existing datasets and algorithms. We believe that our large-scale simulated dataset and the generating pipeline would be a helpful benchmark for future research on F-formation detection, especially on modern data-driven methods. While not as superior as existing algorithms, the end-to-end model would be a great baseline for future works of detecting conversational groups directly from RGB images. This genre of model would significantly reduce the complexity of the annotation process.

## 3. F2SD DATASET

In this section, we describe our proposed approach to generate the simulation dataset. Our pipeline contains two steps: create a 2D map from the SALSA dataset, then use GTA to create the simulation dataset. The pipeline is described in Figure. 2:

1. Body pose, head pose and 2D location information from SALSA are used to randomly synthesize *n* groups in a 2D map. The algorithm is described in Section 3.2. 2D augmentations such as group rotating and translating are also involved implicitly and explicitly in this step through configuration parameters.

2. 2D maps in the first step are used to generate 3D images using a custom modified version of MTA-Mod, a GTA plugin. The in-game camera is randomly located in a suitable range to collect the final images. Inappropriate in-game human models and scenes are manually filtered, resulting in 751 human models and 375 background scene locations. To prevent data leakage, 301 scene locations and 527 human models are used for the training set, and the remaining 74 scene locations and 224 human models are used for the test set.

### 3.1 Parameter

We rely on some configuration parameters to produce a diverse dataset through the two above steps. Through these parameters, we can manage the difficulty level of the dataset and make the dataset more realistic. We divide into two types of parameters: *Explicit parameter* - set up in the intermediate steps, such as the number of groups, group rotation augmentation, etc., and *Implicit parameter* - that people can insight information through simulated images, these parameters are the result of explicit parameter configuration.

**Explicit parameters**

In the simulation procedure, we use four explicit parameters to control the difficulty level of the dataset, which can be manually changed. The explicit parameters are

• *Group rotation*: The range of this parameter is [0º, 360º], with each group being chosen from SALSA. We can rotate a group randomly around its center with this parameter.

• *Group translation*: In an image, locating groups suitably make the image more realistic. An optimization algorithm (described in Section.3.2) is used to locate the position of the group based on the original group arrangement of the actual SALSA dataset.

• *Number of groups*: We randomly choose the number of groups in the range [3, 10]. We observe that the more groups are presented in the image, the more complex input a group detection algorithm will face.

• *Camera configuration*: In the 3D simulation step, we must set up the camera in GTA to collect RGB images. We determine three parameters in this configuration: distance from the camera to the centre of groups in the image, camera height, and camera rotation angle.

**Implicit parameters**

The above explicit parameter permits management of the difficulty level and reality level. Implicit parameters give us insight into the dataset with individual and group occlusion levels. For demonstration, consider the individual occlusion rate, which is the proportion of overlapped regions of a given bounding box, and the group occlusion rate, which is the proportion of the overlapped region on all bounding boxes in a given group. It turns out that higher occlusion rates imposed by our parameters imply a more complex dataset, leading to a performance drop. Therefore, we can manage the difficulty of the dataset via adjustments of implicit and explicit parameters.

## 3.2 2D generation algorithm

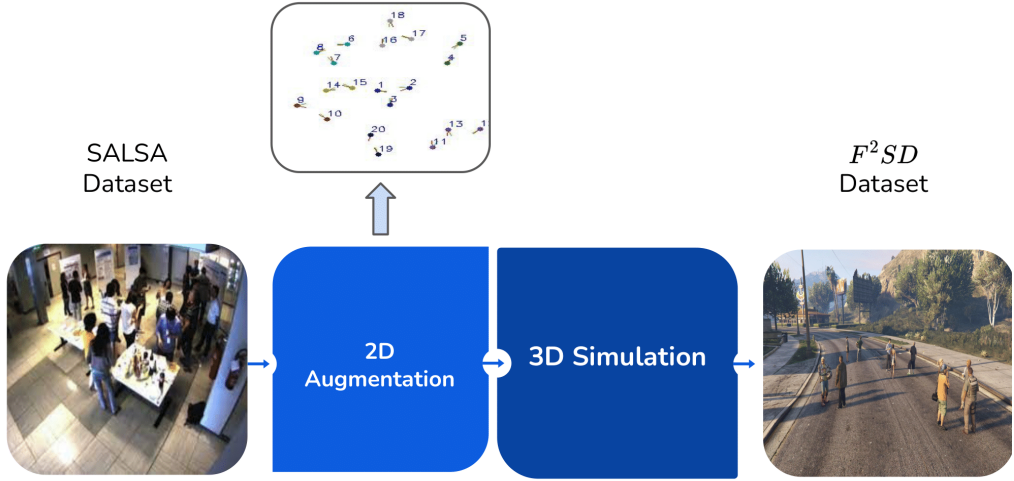

Figure 2: Our simulation generating pipeline.

The above explicit parameter permits management of the difficulty level and reality level. Implicit parameters give us insight into the dataset with individual and group occlusion levels. For demonstration, consider the individual occlusion rate, which is the proportion of overlapped regions of a given bounding box, and the group occlusion rate, which is the proportion of the overlapped region on all bounding boxes in a given group. It turns out that higher occlusion rates imposed by our parameters imply a more complex dataset, leading to a performance drop. Therefore, we can manage the difficulty of the dataset via adjustments of implicit and explicit parameters.

Prior to 3D data generation, 2D scene geometry, i.e., a description of how people in the scene are located and oriented, is required. To ease the process of scene geometry simulation, we exploit realistic datasets like SALSA as prior information about regular scene geometry. Concretely, a pool of group geometries, i.e., a description of how people in a group are located and oriented, is taken from realistic datasets. Our algorithm simulates each scene geometry by iteratively sampling a group geometry from the pool and locating it to the scene geometry so that the resulting scene geometry preserves the regular group-to-group distance as in realistic datasets.

Consider a scene with $N - 1$ groups $G(1), \ldots, G(N - 1)$, we want to add a new group $G(N)$ sampled from the group geometry pool to that scene. The problem of locating $G(N)$ is formulated as an optimization problem, with the loss function indicating two objectives:

- Objective 1: $G(N)$ should not be too close to the $N - 1$ existing groups, as given in equation (3) below.
- Objective 2: $G(N)$ should not be too far from the $N - 1$ existing groups, as given in equation (4) below.

In order to mathematically model the objectives, we represent each group $G(i)$ as a Gaussian $N(\mu^{(i)}, \Sigma^{(i)})$ where $\mu^{(i)}$ is the group centre and $\Sigma^{(i)}$ is the covariance matrix of group members' locations. Following this representation, we define two functions: inter-group distance and group variance along a particular direction.

Concretely, the inter-group distance between $G(i)$ and $G(j)$ is given as

$$d(G(i), G(j)) = ||\mu(i) - \mu(j)||^2 \qquad (1)$$

and the group variance of $G(i)$ in the direction to $G(j)$ is formulated as [15]

$$\sigma(G^{(i)}; G^{(j)}) = \sqrt{e(\mu^{(j)} - \mu^{(i)}) \cdot \Sigma^{(i)} \cdot e(\mu^{(j)} - \mu^{(i)})^T} \qquad (2)$$

where $e(x) = \frac{x}{||x||_2}$ denotes the normalization function indicating the direction of $\mu(j) - \mu(i)$. For convenience, we abbreviate $d(G^{(i)}, G^{(j)})$ as $d(i, j)$ and $\sigma(G^{(i)}; G^{(j)})$ as $\sigma(i; j)$

### 3.2.1 Objective 1

Given a new group $G^{(N)}$ and an existing group $G^{(i)}$, we model the first objective by requiring that $\sigma(N; i)$ and $\sigma(i; N)$ is small enough compared to $d(N, i)$, i.e.

$$\begin{aligned} r_d \cdot \sigma(N; i) + \beta &\le d(N, i) \\ r_d \cdot \sigma(i; N) + \beta &\le d(N, i) \end{aligned} \qquad (3)$$

where $\beta$ is a margin parameter as in Triplet loss[16], and $r_d = E_{i,j}(\frac{d(i,j)}{\sigma(i;j)})$ is the average distance-to-variance ratio estimated from all group pairs (within the same image) of the SALSA dataset.

### 3.2.2 Objective 2

The second objective is modelled by constraining $d(N, i)$ below a certain threshold $\gamma$, i.e.

$$d(N, i) \le \gamma \qquad (4)$$

### 3.2.3 Loss function

We then define the loss function for the new group $G^{(N)}$ w.r.t an existing group $G^{(i)}$ as

$$L^{(i)}(G^{(N)}) = max\{r_d \cdot \sigma(N; i) + \beta - d(N, i), 0\}$$

$$+ max\{r_d \cdot \sigma(i; N) + \beta - d(N, i), 0\} \qquad (5)$$

$$+ max\{\gamma - d(N, i), 0\}$$

$L^{(i)}(G^{(N)})$ may push the new group too far from existing groups, by making $\sigma(N; i)$ and $\sigma(i; N)$ too small relative to $d(N, i)$. To prevent this issue, we introduce a regularizer. The overall loss function for $G^{(N)}$ is hence given as

$$L(G^{(N)}) = \frac{1}{N-1} \sum_{i=1}^{N-1} L^{(i)}(G^{(N)}) + \theta R(G^{(N)}) \qquad (6)$$

where $\theta$ is the regularization rate. By optimizing $\mu(N)$ to minimize $L(G^{(N)})$, we would be able to properly locate the new group $G^{(N)}$ in the scene with $N - 1$ existing groups $G^{(1)}, \ldots, G^{(N-1)}$. In this work, we exploit Adam optimizer [17] with learning rate $5 \cdot 10^{-2}$. The two parameters $\beta$ and $\gamma$ are set to $0.4$ and $2.0$ respectively.

## 3.3 Dataset statistics

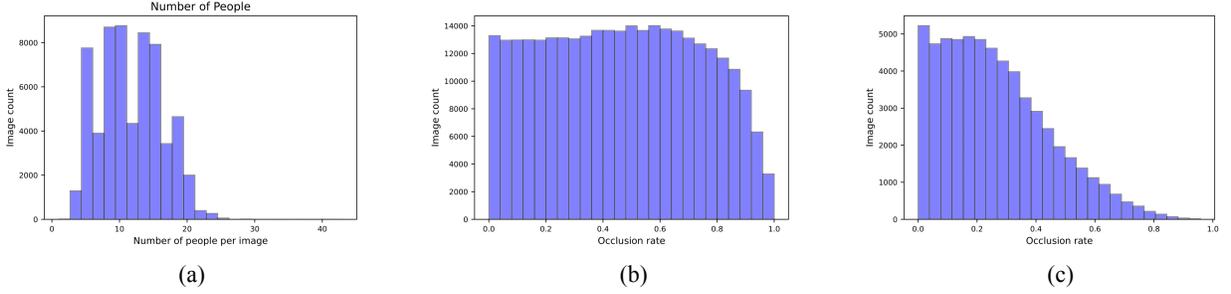

Figure 3: Statistics on our simulation dataset
(a) number of people per image distribution, (b) individual occlusion distribution, (c) group occlusion distribution.

In this section, we summarize the statistics of the dataset. The *F2SD* dataset contains a total of 61,967 images containing 732,248 people instances. Each image has a set of groups, including people's IDs in each group. We split the *F2SD* dataset into a training split and a test split. The number of images, annotated people instances and number groups in each of these splits is tabulated in *Table. 1*. *Figure. 3 (a)* shows the distribution of the number of people in each image in our simulation dataset. On average, these have 3 groups and 12 people per image. Besides, we also annotated the bounding box for the SALSA dataset. In the SALSA dataset, *Figure 5 (a)* visualizes the distribution of the number of groups in one image, with a maximum of 7 groups, we do not count groups having a number of people equal to 1.

Each image in SALSA has 18 people participating in a poster session or a coffee break. *Figure 3 (b), (c)* shows the occlusion distribution of individual and group in our dataset.

## 3.4 Implicit parameter

### 3.4.1 Individual occlusion

The individual non-occlusion rate of a particular bounding box is given as the area of the non-occluded proportion of that box, divided by the total area of the box. Concretely, considering a bounding box $B^{(k)}$ and other boxes occluding it $B^{(k1)}, B^{(k2)}, \ldots, B^{(k_{n_k})}$, the non-occlusion rate of $B^{(k)}$ is given as

$$NO_I(B^{(k)}) = \frac{|R(B^{(k)}) \setminus \bigcup_{i=1}^{n_k} R(B^{(k_i)})|}{|R(B^{(k)})|} \quad (7)$$

where $R(B)$ denotes the set of pixels covered by a box $B$, and $|\cdot|$ denotes set cardinality operator. The individual occlusion rate is then formulated as

$$O_I(B^{(k)}) = 1 - NO_I(B^{(k)}) \quad (8)$$

### 3.4.2 Group occlusion

To define group occlusion, we define the non-occluded part of a group as the union of non-occluded parts of all bounding boxes in that group. The total area of a group is defined similarly as the area of the union of all bounding boxes in the group. The group non-occlusion rate is then given as the area of the non-occluded proportion of that group, divided by the total area of the group. Formally, the group non-occlusion rate of a group $G(i)$ consisting of bounding boxes $B_1^{(i)}, B_2^{(i)}, \ldots, B_{n_i}^{(i)}$ is given as

$$\frac{|\bigcup_{j=1}^{n_i} NO_I(B_j^{(i)})|}{|\bigcup_{j=1}^{n} R(B_j^{(i)})|} \qquad (9)$$

## 4. END-TO-END GROUP DETECTION MODEL

In this section, we present the technique of our proposed end-to-end baseline with problem formulation and mode architecture.

| # | train | test | all |
|---|---|---|---|
| Number of image | 40,234 | 21,733 | 61,967 |
| Number of people | 475,332 | 256,916 | 732,248 |

Table 1: Statistics of various splits of *F2SD*

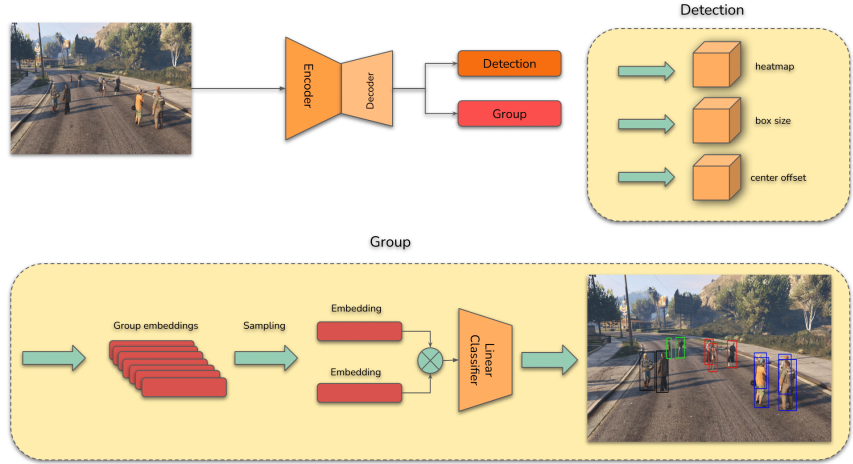

Figure 4: The proposed end-to-end architecture

### 4.1 Problem formulation

We consider group detection an end-to-end problem where we can apply a deep neural network on an RGB image to simultaneously produce bounding box and group information. Our model is built primarily based on Fairmot [18]. Fairmot is a one-stage tracking framework composed of three components: a backbone model for extracting features from images and two head serving detection and re-id tasks. In our model, instead of using a re-id head, we replace it with a group branch responsible for extracting group features and classifying whether two persons belong to one group. From the output of the group head, we will build a connected components graph to solve the group detection problem.

## 4.2 Model architecture

Fairmot [18] adopts ResNet-34 as its backbone in order to achieve a good balance of accuracy and speed. The authors use an enhanced version of Deep Lauer Aggregation (DLA) [19], which has more skip connections between low-level and high-level features. Denote size of input image as $H_{image} \times W_{image}$, then the output feature map has the shape of $C \times H \times W$ where $H = H_{image}/4$ and $W = W_{image}/4$.

We utilize the detection branch, heatmap head and box offset, size heads from Fairmot. In particular, the dimension of the heatmap is $1 \times H \times W$. For each GT box $b^i = (x_1^i, y_1^i, x_2^i, y_2^i)$ in the image, the author computes the object center $c_x^i, c_y^i$ as $c_x^i = \frac{x_1^i + x_2^i}{2}$ and $c_y^i = \frac{y_1^i + y_2^i}{2}$, respectively. Then its location on the feature map is obtained by dividing the stride $(\tilde{c}_x^i, \tilde{c}_y^i) = (\lfloor \frac{c_x^i}{4} \rfloor, \lfloor \frac{c_y^i}{4} \rfloor)$. We use exactly the loss function of heatmap and box: $L_{heat}$ and $L_{box}$ from the FairMot framework. The overall architecture is illustrated in Figure 4.

**Group embedding branch**

The group embedding branch aims to generate features that distinguish objects in different groups. To achieve this goal, we utilize a feature map as $E \in R^{128 \times H \times W}$ as the Re-ID branch [18]. The group feature $E_{x,y} \in R^{128}$ of an object centred at $(x, y)$ can be extracted from the feature map.

In the training phase, we sample positive and negative pairs and combine these embeddings to the classification task. All object instances in the same group of the training set are treated as the same class. All object instances in the different groups in the same image are treated as different classes. For each pair of GT box $b^i = (x_1^i, y_1^i, x_2^i, y_2^i)$ and $b^{i'} = (x_1^{i'}, y_1^{i'}, x_2^{i'}, y_2^{i'})$ in the image, we obtain the object centre on the heatmap $(\tilde{c}_x^i, \tilde{c}_y^i)$ and $(\tilde{c}_x^{i'}, \tilde{c}_y^{i'})$. We extract the group feature vector $E_{\tilde{c}_x^i, \tilde{c}_y^i}$ and $E_{\tilde{c}_x^{i'}, \tilde{c}_y^{i'}}$, then combine two group embeddings using concatenation operation, sum operation or average operation. For the group classification model, we use a fully connected layer and a sigmoid operation to map this combined embedding to a binary output 0, 1. We use binary entropy classification for loss of group branch - $L_{group}$. We jointly train the detection and the group embedding branches by adding the losses $L_{heat}, L_{box}, L_{group}$, the framework with two models is trained end-to-end:

$$L_{detection} = L_{heat} + L_{box} \qquad (10)$$

$$L_{total} = w1 \cdot L_{detection} + w2 \cdot L_{group} \qquad (11)$$

where $w1$ and $w2$ are hyperparameters that regulate detection loss and group loss, respectively.

In the inference phase, we get a bounding box of people and embedding for each bounding box. Then, we consider $n$ detected people as a graph having $n$ vertices, where edges are determined by the group classification model with the input being the combination of pair embeddings. Finally, we extract connected components in the graph to find groups.

## 5. EXPERIMENT

We summarise our results on our simulation dataset in this section, the metric F1 scores for $T = 2/3$ (i.e., predictions with a 2/3 match to ground truth are considered correct as described by Setti et al. [14]) and $T = 1$. We show the result of three baselines, which are three combination approaches: average, sum and concatenate operation. Results in *Table 2* show the performance of three baseline models on our simulation dataset with test split. Our model achieves a high F1 score on the overall test split. Because the SALSA dataset lacks the number of images and its images are low-quality, we cannot experiment with our model directly on the SALSA dataset. We extract nearly one thousand images from the test set (named $F2SD_{occlusion}$) and experiment on this subset, with the result being shown in *Table 3*. The F1 score on this subset is lower than average because this subset has a higher level of individual and group occlusion. By manually annotating bounding boxes for the SALSA dataset, we observe that its individual and group occlusion levels are high. *Figure 5 (a), (b)* show the occlusion level distribution on SALSA. These occlusion statistics are computed from our bounding box annotation. The dataset has a very high individual and group occlusion level, and our $F2SD_{occlusion}$ also has a similar occlusion level with it.

| Experiment | F1@2/3 | F1@1 | mAP detection |
|---|---|---|---|
| Concatenate operation | **0.842** | **0.823** | 0.993 |
| Average operation | 0.825 | 0.800 | 0.993 |
| Sum operation | 0.827 | 0.804 | 0.993 |

Table 2: Experiment on our test simulation dataset

| Experiment | F1@2/3 | F1@1 | mAP detection |
|---|---|---|---|
| Concatenate operation | **0.612** | **0.575** | 0.986 |
| Average operation | 0.538 | 0.503 | 0.986 |
| Sum operation | 0.560 | 0.525 | 0.986 |

Table 3: Experiment on our subset from the test set having similar individual and group occlusion with the SALSA ($F2SD_{occlusion}$)

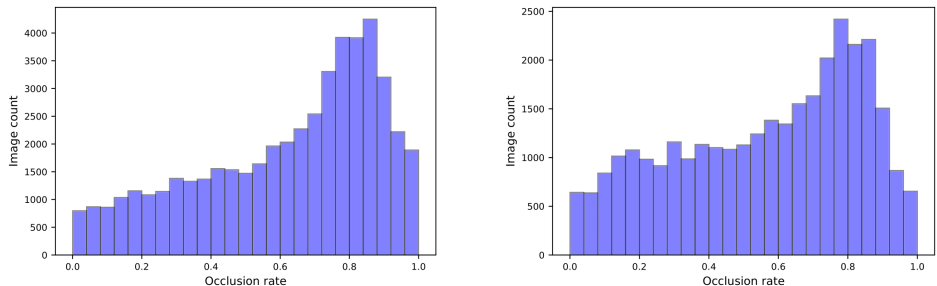

Figure 5: Statistics on the SALSA dataset based on our bounding box annotation with number of people per image is fixed at 18 (a) individual occlusion distribution, (b) group occlusion distribution.

| Graph-cut rate | Link threshold | Original | Finetuning |
|---|---|---|---|
| 0.2 | 0.5 | 53.13 ± 0.5 | 55.10 ± 0.7 |
| 0.2 | 0.7 | 60.01 ± 0.15 | 62.28 ± 0.15 |
| 0.2 | 0.9 | 68.78 ± 0.3 | 70.54 ± 0.1 |
| 0.3 | 0.9 | 71.93 ± 0.6 | 72.61 ± 0.2 |

Table 4: Experiment on the SALSA dataset following metric **F1@2/3**

*Table 4* shows us that training the end-to-end model with the simulation dataset can improve the F1 score on the SALSA dataset. After fitting the image into the model, we use a graph-based method as post-processing to cluster the groups. Graph-cut rate and Link threshold are two hyperparameters for this algorithm. Our proposed model training on the SALSA dataset with the K-fold scheme is the original result. We use $K = 5$, and in each fold, we divide 80% for the training set and 20% for the test set. Fine-tuning is the result that we first train our model with the simulation dataset, then fine-tune it on the SALSA dataset. As we can see, our simulation dataset can improve the F1 score on the SALSA dataset. Our best achievement on the SALSA dataset is nearly 72.61 F1@2/3, which is comparable to 81.2 from [9].

## 6. CONCLUSION

Existing works on F-Formation detection are limited in the data quantity or methods. We presented a novel pipeline to generate data for f-formation detection in this work. We create the first large-scale simulated dataset using Grand Theft Auto V and the pipeline to generate it as a new contribution. We use implicit and explicit configurations to manage the complexity and practical level of the simulation dataset. We also propose a novel end-to-end baseline method to directly capture group information from RGB images, achieving approximately 0.82 in F1 score on the overall test set. We also experiment on subset sampling from the test set having occlusion levels similar to the SALSA dataset. Our model achieves a 0.57 F1 score on the subset having similar individual and group occlusion with the SALSA dataset. In the future, we will improve our neural network on the actual situation using the simulation dataset. We will also create a large-scale simulation dataset that can cover all real-life situations such as indoor and outdoor group activities.

## ACKNOWLEDGMENTS

The authors would like to thank the VNU University of Engineering and Technology, and Dai Nippon Printing Co., Ltd. for supporting this study

## REFERENCES

[1] Kendon, A., "Spacing and orientation in co-present interaction," in [Development of multimodal interfaces: Active listening and synchrony ], 1–15, Springer (2010).
[2] Mulford, C. L., "Erving goffman," behavior in public places. notes on the social organization of gatherings"(book review)," Sociological Quarterly 6(2), 181 (1965).
[3] Kendon, A., [Conducting interaction: Patterns of behavior in focused encounters ], vol. 7, CUP Archive (1990).
[4] Alameda-Pineda, X., Subramanian, R., Ricci, E., Lanz, O., and Sebe, N., "Salsa: A multimodal dataset for the automated analysis of free-standing social interactions," in [Group and Crowd Behavior for Computer Vision ], 321–340, Elsevier (2017).

[5] Zen, G., Lepri, B., Ricci, E., and Lanz, O., "Space speaks: Towards socially and personality aware visual surveillance," in [Proceedings of the 1st ACM International Workshop on Multimodal Pervasive Video Analysis], MPVA '10, 37–42, Association for Computing Machinery, New York, NY, USA (2010).

[6] Setti, F., Lanz, O., Ferrario, R., Murino, V., and Cristani, M., "Multi-scale f-formation discovery for group detection," in [2013 IEEE International Conference on Image Processing], 3547–3551 (2013).

[7] Hung, H. and Kröse, B., "Detecting f-formations as dominant sets," in [Proceedings of the 13th international conference on multimodal interfaces], 231–238 (2011).

[8] Swofford, M., Peruzzi, J. C., Vázquez, M., Martín-Martín, R., and Savarese, S., "DANTE: deep affinity network for clustering conversational interactants," CoRR abs/1907.12910 (2019).

[9] Hedayati, H., Muehlbradt, A., Szafir, D. J., and Andrist, S., "Reform: Recognizing f-formations for social robots," (2020).

[10] Krishna, S., Kiselev, A., and Loutfi, A., "Towards a method to detect f-formations in real-time to enable social robots to join groups," in [Towards a Method to Detect F-formations in Real-Time to Enable Social Robots to Join Groups :], Umeå University (2017).

[11] Hung, H. and Kröse, B., "Detecting f-formations as dominant sets," in [Proceedings of the 13th international conference on multimodal interfaces], 231–238 (2011).

[12] Cristani, M., Bazzani, L., Paggetti, G., Fossati, A., Tosato, D., Del Bue, A., Menegaz, G., and Murino, V., "Social interaction discovery by statistical analysis of f-formations.," in [BMVC], 2, 4, Citeseer (2011).

[13] Bazzani, L., Cristani, M., Tosato, D., Farenzena, M., Paggetti, G., Menegaz, G., and Murino, V., "Social interactions by visual focus of attention in a three-dimensional environment," Expert Systems 30(2), 115–127 (2013).

[14] Setti, F., Russell, C., Bassetti, C., and Cristani, M., "F-formation detection: Individuating free-standing conversational groups in images," PloS One 10(5), e0123783 (2015).

[15] Shalizi, C., "Advanced methods of data analysis," (Spring 2012).

[16] Schroff, F., Kalenichenko, D., and Philbin, J., "Facenet: A unified embedding for face recognition and clustering," in [Proceedings of the IEEE conference on computer vision and pattern recognition], 815–823 (2015).

[17] Kingma, D. P. and Ba, J., "Adam: A method for stochastic optimization," (2014).

[18] Zhang, Y., Wang, C., Wang, X., Zeng, W., and Liu, W., "Fairmot: On the fairness of detection and re-identification in multiple object tracking," International Journal of Computer Vision 129(11), 3069–3087 (2021).

[19] Zhou, X., Wang, D., and Krähenbühl, P., "Objects as points," arXiv preprint arXiv:1904.07850 (2019).